\documentclass[letterpaper, 10 pt, conference]{ieeeconf} 

\IEEEoverridecommandlockouts
\usepackage{graphicx} 
\usepackage{amsmath} 
\usepackage{amssymb}  
\usepackage{gensymb}
\usepackage{cite}
\usepackage{multirow}
\usepackage[hyphens]{url}
\usepackage{hyperref}

\begin{document}

\title{\LARGE \bf Toward Ergonomic Risk Prediction via Segmentation of Indoor Object Manipulation Actions Using Spatiotemporal Convolutional Networks}

\author{Behnoosh Parsa$^{1}$, Ekta U. Samani$^{1}$, Rose Hendrix$^{1}$, Cameron Devine$^{1}$, Shashi M. Singh$^{2}$, Santosh Devasia$^{1}$,\\ and Ashis G. Banerjee$^{3}$
\thanks{This work was supported in part by a generous gift from Amazon Robotics.}
\thanks{$^{1}$B. Parsa, E. Samani,  R. Hendrix, C. Devine, and S. Devasia are with the Department of Mechanical Engineering, University of Washington, Seattle, WA 98195, USA, {\tt\footnotesize behnoosh,ektas,rmhend,camdev, devasia@uw.edu}}%
\thanks{$^{2}$S. M. Singh is with the Department of Mechanical Engineering, Indian Institute of Technology Gandhinagar, Palaj 382355, Gujarat, India, {\tt\footnotesize shashi.singh@iitgn.ac.in}}%
\thanks{$^{3}$A. G. Banerjee is with the Department of Industrial \& Systems Engineering and the Department of Mechanical Engineering, University of Washington, Seattle, WA 98195, USA, {\tt\footnotesize ashisb@uw.edu}}%
}

	\maketitle
	
	\begin{abstract}
	Automated real-time prediction of the ergonomic risks of manipulating objects is a key unsolved challenge in developing effective human-robot collaboration systems for logistics and manufacturing applications. We present a foundational paradigm to address this challenge by formulating the problem as one of action segmentation from RGB-D camera videos. Spatial features are first learned using a deep convolutional model from the video frames, which are then fed sequentially to temporal convolutional networks to semantically segment the frames into a hierarchy of actions, which are either ergonomically safe, require monitoring, or need immediate attention. For performance evaluation, in addition to an open-source kitchen dataset, we collected a new dataset comprising twenty individuals picking up and placing objects of varying weights to and from cabinet and table locations at various heights. Results show very high (87-94)\% F1 overlap scores among the ground truth and predicted frame labels for videos lasting over two minutes and consisting of a large number of actions. 
	\end{abstract}

    	
	\section{Introduction}
	One of the key considerations for viable human-robot collaboration (HRC) in industrial settings is \textit{safety}. This consideration is particularly important when a robot operates in close proximity to humans and assists them with certain tasks in increasingly automated factories and warehouses. Therefore, it is not surprising that a lot of effort has gone into identifying and implementing suitable HRC safety measures \cite{zanchettin2016safety}. Typically, the efforts focus on designing collaborative workspaces to minimize interferences between human and robot activities \cite{michalos2015design}, installing redundant safety protocols and emergency robot activity termination mechanisms through multiple sensors \cite{michalos2015design}, and developing both predictive and reactive collision avoidance strategies \cite{robla2017working}. These efforts have resulted in the expanded acceptance and use of industrial robots, both mobile and stationary, leading to increased operational flexibility, productivity, and quality \cite{maurtua2017human}.

	A key factor in achieving safe HRC is accurate robotic perception of humans actions and their potential risks. Specifically, perceiving (assessing) the ergonomic risks of human actions is an extremely important topic that has not received much attention until recently \cite{golabchi2015automated, li2019automated}. Unlike other commonly considered safety measures, a lack of ergonomic safety does not lead to immediate injury concerns or fatality risks. It, however, causes or increases the likelihood of causing longterm injuries in the form of musculoskeletal disorders (MSDs) \cite{helander2005guide}. According to a recent report by the U.S. Bureau of Labor Statistics, there were 349,050 reported cases of MSDs in 2016 just in the U.S. \cite{usbureauoflaborstatistics2018}, leading to tens of billions of dollars in healthcare costs. 
	
	Most organizations use conventional ergonomic risk assessment methods, which are based on observations and self-reports, making them error-prone, time consuming, and labor-intensive \cite{spielholz2001comparison}. More recently, researchers have started exploring alternative sensor-based automated assessment methods. For example, Li et al. \cite{li2011computer} used distributed surveillance cameras and body-mounted motion sensors for this purpose. Shafti et al. \cite{Shafti2019} used an RGB-D camera to extract the skeletal information of the arm and understand the safe range of arm motions during welding. Kim et al. \cite{Kim2019} introduced a reconfigurable HRC workstation to monitor and adjust the ergonomic risks of working with power tools in real time using a stereovision camera. 
    
    From a methodological perspective, deep learning has become popular in assessing the risks of performing occupational tasks, especially in the construction industry \cite{fang2018falls}. Outside of the construction sector, Abobakr et al. \cite{abobakr2017kinect} employed deep residual convolutional networks (CNNs) to predict the joint angles of manufacturing workers from individual camera depth images. Mehrizi et al. \cite{mehrizi2018toward} proposed a multi-view based deep perceptron approach for markerless 3D pose estimation in the context of object lifting tasks. While all these works present useful advances and report promising performances, they do not provide a general-purpose framework to predict the ergonomic risks for any representative set of object manipulation tasks commonly performed in the industry.
    
    Here, we present a first of its kind \textit{end-to-end} deep learning system for ergonomic risk assessment during indoor object manipulation using camera videos. Our learning system is based on \textit{action segmentation}, where an action class (with a corresponding risk label) is predicted for every video frame. Representative works on this topic include that by Fathi et al. \cite{fathi2013modeling}, who showed that state changes at the start and end frames of actions provided good segmentation performance. Kuehne et al. \cite{kuehne2016end} used reduced Fisher vectors for visual (spatial) representation of every frame, which were then fitted to Gaussian mixture models. Huang et al. \cite{huang2016connectionist} presented a temporal classification framework in the case of weakly supervised action labeling. Ghosh et al. \cite{ghosh2018stacked} recently developed a graph-based spatiotemporal CNN to exploit environmental cues for better segmentation. Along similar lines, our method uses a combination of spatial and temporal CNNs to achieve good segmentation performance.
 
 In addition, we present a new benchmark dataset, called the University of Washington Indoor Object Manipulation (UW IOM) dataset, for vision-based ergonomic risk assessment studies. Given an acceptable ergonomic risk model, we then show that our end-to-end system satisfactorily predicts the risks of actions in test videos.  The goal of our system, therefore, is to enable the collaborative robots to accurately detect the risky manipulation actions so that they can perform these actions, allowing the humans to instead engage in supervisory control or cognitively challenging tasks.

	\section{Ergonomic Risk Assessment Model}
	We use a well-established ergonomic model, known as the rapid entire body assessment (REBA) model \cite{hignett2004rapid}, which is popularly used in the industry. The REBA model assigns scores to the human poses, within a range of 1-15, on a frame-by-frame basis by accounting for the joints motions and angles, load conditions, and activity repetitions. An action with an overall score of less than 3 is labeled as ergonomically safe, a score between 3-7 is deemed to be medium risk that requires monitoring, and every other action is considered high risk that needs attention.
	
	Skeletal information for the TUM Kitchen dataset \cite{tenorth2009tum} is available in the bvh (Biovision Hierarchy) file format. We use the bvh parser from the MoCap Toolbox \cite{burger2013mocap} in MATLAB to read this information as the XYZ coordinates of thirty three markers (joints and end sites) corresponding to every frame. For the UW IOM dataset, the positions of twenty five different joints are recorded directly in the global coordinate system for each frame using the Kinect sensor with the help of a toolbox \cite{terven2016kin2} that links Kinect and MATLAB. For every frame, the vectors corresponding to different body segments such as fore-arm, upper-arm, leg, thigh, lower half spine, upper half spine, and so on, are computed. The extension, flexion, and abduction (as applicable) of the various body segments are computed by taking the projection of the two body segment vectors that constitute the angle on the plane of motion. These angles of extension, flexion, and abduction are used to assign the trunk, neck, leg, upper arm, lower arm, and wrist scores \cite{hignett2004rapid}.
	
	We define three different thresholds as a part of our implementation, namely, zero threshold, binary threshold, and abduction threshold. Zero threshold is used for trunk bending, such that any trunk bending angle less than this value is regarded as no bending. Binary threshold is defined to answer whether the trunk is twisted and/or side flexed. Trunk twisting and trunk side flexion less than this value are ignored. Abduction threshold, though similar to the binary threshold, is separately defined for shoulder abduction considering the considerably larger allowable range of abduction (about $150^{\circ}$) as against a smaller allowable range of trunk twisting. Due to the non-availability of rotation information of the neck, we assume that the neck is twisted when the trunk is twisted, which is not entirely unreasonable.
	The nature of the performed actions does not involve arm rotations, and they are ignored while computing the upper arm score.
	
	The computed trunk, neck, leg, upper arm, lower arm, and wrist scores are used to assign the REBA score on a frame-by-frame basis using lookup tables \cite{hignett2004rapid}. The REBA scores assigned for each frame are then aggregated over all the actions and participants, so that we have a constant risk score for every frame that corresponds to a particular action. This aggregated value is also considered as the final REBA score for that particular action.

	\section{Deep Learning Models}
	\subsection{Spatial Features Extraction}
	
	We adapt two variants of VGG16 convolutional neural network models \cite{simonyan2014very} for spatial feature extraction. The first model is based on the VGG16 model that is pre-trained on the ImageNet database \cite{deng2009imagenet}. The second model involves fine-tuning the last two convolutional layers of the VGG16 base that is pretrained on ImageNet. In both the models, the flattened output of the last convolutional layer is connected to a fully connected layer with a drop-out of 0.5 and then fed into a classifier. We always use rectified linear units (ReLU) and softmax as the activation functions, and Adam \cite{kingma2014adam} as the optimizer.
	
	We also use a simplified form of the pose-based CNN (P-CNN) features \cite{cheron2015p} that only consider the full images and not the different image patches. Optical flow \cite{brox2004high} is first computed for each consecutive pair of RGB datasets, and the flow map is stored as an image \cite{gkioxari2015finding}. A motion network, introduced in \cite{gkioxari2015finding}, containing five convolutional and three fully-connected layers, is then used to extract frame descriptors for all the optical flow images. Subsequently, the VGG-f network \cite{chatfield2014return}, pre-trained on ImageNet, is used to extract another set of frame descriptors for all the RGB images. The VGG-f network also contains five convolutional and three fully connected layers. The two sets of frame descriptors are put together as arrays in the same sequence as that of the video frames to construct motion-based and appearance-based video descriptors, respectively. The appearance and motion-based video descriptors are then normalized and concatenated to form the final video descriptor (spatial features).
	
	\subsection{Video Segmentation Methods}
	We use two kinds of temporal convolutional networks (TCNs), both of which use encoder-decoder architectures to capture long-range temporal patterns in videos \cite{lea2017temporal}. In the first network, referred as the encoder decoder-TCN, or ED-TCN, a hierarchy of temporal convolutions, pooling, and upsampling layers is used.  The network does not have a large number of layers, but each layer includes a set of long convolutional filters. We use the ReLU activation function and a categorical cross-entropy loss function with RMSProp \cite{csc321} as the optimizer. In the second network, termed as dilated-TCN, or D-TCN, dilated upsampling and skip connections are added between the layers. We use the gated activation function as it is inspired by the WaveNet \cite{van2016wavenet} and Adam optimizer. We also use two other segmentation methods for comparison purposes. The first method is bidirectional long short term memory, or Bi-LSTM\cite{graves2005bidirectional}, a recurrent neural network commonly used for analyzing sequential data streams. The second method is support vector machine, or SVM, which is extremely popular for any kind of classification problem.

	\subsection{Video Segmentation Performance Metrics}
	In addition to frame-based accuracy, which is the percentage of frames labeled correctly for the related sequence as compared to the ground truth (manually annotated), we report edit-score and F1 overlap score to evaluate the performance of the various methods. The edit-score \cite{lea2016learning} measures the correctness of Levenshtein distance to the segmented predictions. The F1 overlap score \cite{lea2016learning}, combines classification precision and recall to reduce the sensitivity of the predictions to minor temporal shifts between the predicted and ground truth values, as such shifts might be caused by subjective variabilities among the annotators.
	
	\section{System Architecture and Datasets}
	\subsection{System Architecture}
\begin{figure*}
	    \centering
        \includegraphics[width=0.75\textwidth]{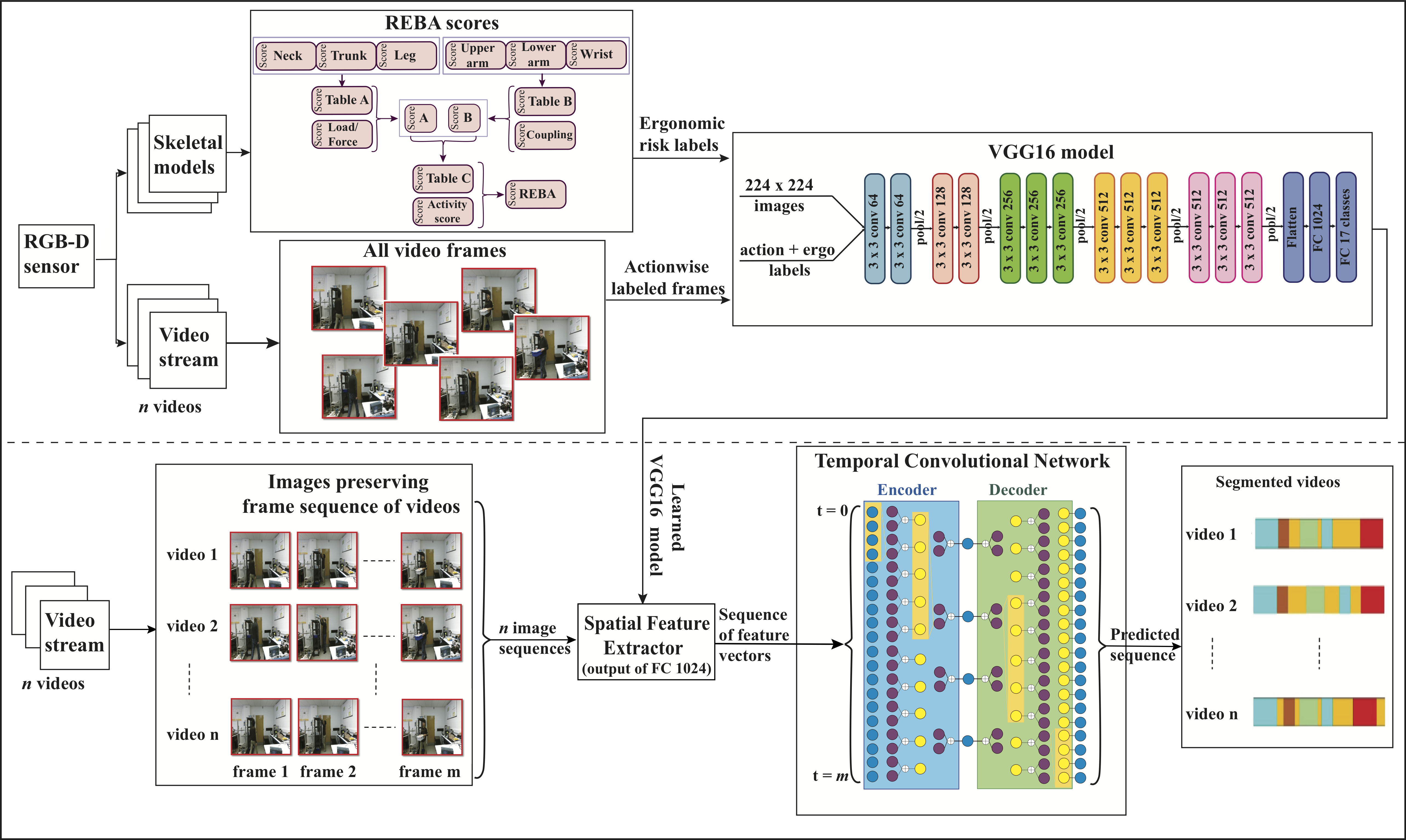}
        \caption{End-to-end ergonomic risk prediction system}
        \label{pipeline}
    \end{figure*}

	We develop an end-to-end automated ergonomic risk prediction system as shown in Fig. \ref{pipeline}. The Figure shows that the prediction works in two stages. In the first stage (top half of the Figure), which only needs to be done once for a given dataset, ergonomic risk labels are computed for each object manipulation action class based on the skeletal models extracted from the RGB-D camera videos. Simultaneously, the videos are annotated carefully to assign an action label to each and every frame. These two types of labels are then used to learn a modified VGG16 model for the entire set of available videos. In the second stage (bottom half of the Figure), during training, the exact sequence of video frames is fed to the learned VGG16 model to extract useful spatial features. The array of extracted features is then fed in the same order to train a TCN on how to segment the videos by identifying the similarities and changes in the features corresponding to actions executions and transitions, respectively. For testing, a similar procedure is followed except that the trained TCN is now employed to segment unlabeled videos into semantically meaningful actions with known ergonomic risk categories. It is possible to replace the VGG16 model by another deep neural network model that also accounts for human motions (e.g., P-CNN) to achieve slightly better segmentation performance at the expense of longer training and prediction times.
	
	\subsection{Datasets}
	
	\subsubsection{TUM Kitchen Dataset}
	The TUM Kitchen dataset \cite{tenorth2009tum}
	consists of nineteen videos, at twenty-five frames per second, taken by four different monocular cameras, numbered from 0 to 3. Each video captures regular actions performed by an individual in a kitchen involving walking, picking up, and placing utensils to and from cabinets, drawers, and tables. The average duration of the videos is about two minutes. The dataset also includes skeletal models of the individual through 3D reconstruction of the camera images. These models are constructed using a markerless full body motion tracking system through hierarchical sampling for Bayesian estimation and layered observation modeling to handle environmental occlusions \cite{bandouch2009tracking}. We categorize the actions into twenty-one classes or labels, where each label follows a two-tier hierarchy with the first tier indicating a motion verb (close, open, pick-up, place, reach, stand, twist, and walk) and the second tier denoting the location (cabinet, drawer) or mode of object manipulation (do not hold, hold with one hand, and hold with both hands).

	\subsubsection{UW IOM Dataset}
\begin{figure}
            \includegraphics[width=0.5\textwidth]{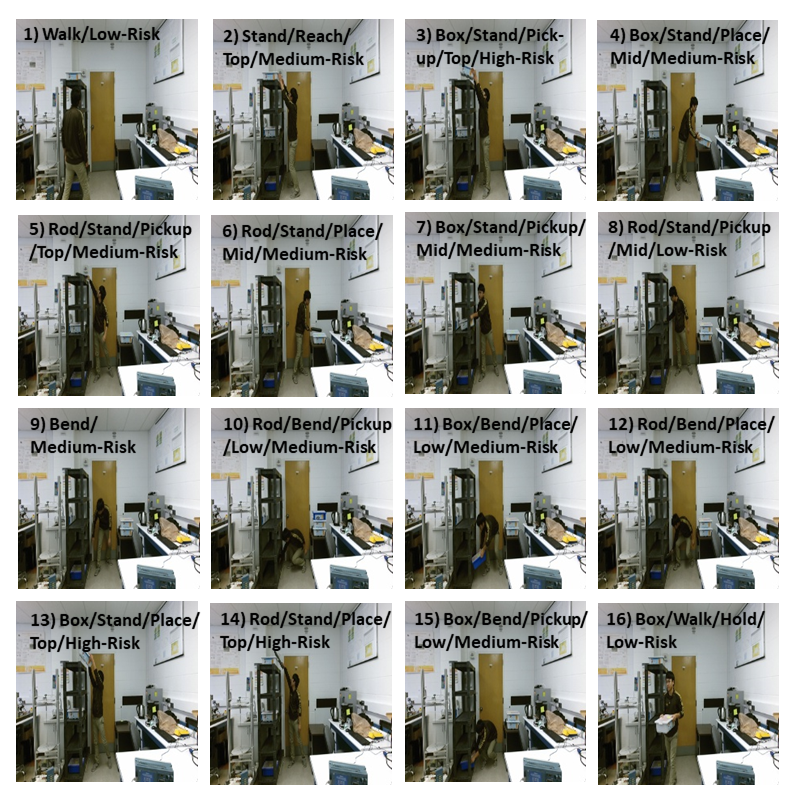}
     
        \caption{Representative video frames depicting actions with different ergonomic risk levels in our own UW IOM dataset}
        \label{actions}
    \end{figure}

	Considering the dearth of suitable videos capturing object manipulation actions involving awkward poses and repetitions, we collected our own dataset using an Institutional Review Board (IRB)-approved study. The dataset comprises videos of twenty participants within the age group of 18-25 years, of which fifteen are males and the remaining five are females. The videos are recorded using a Kinect Sensor for Xbox One at an average rate of twelve frames per second. Each participant carries out the same set of tasks in terms of picking up six objects (three empty boxes and three identical rods) from three different vertical racks, placing them on a table, putting them back on the racks from where they are picked up, and then walking out of the scene carrying the box from the middle rack. The boxes are manipulated with both the hands while the rods are manipulated using only one hand. The above tasks are repeated in the same sequence three times such that the duration of every video is approximately three minutes. We categorize the actions into seventeen labels, where each label follows a four-tier hierarchy. The first tier indicates whether the box or the rod is manipulated, the second tier denotes human motion (walk, stand, and bend), the third tier captures the type of object manipulation if applicable (reach, pick-up, place, and hold), and the fourth tier represents the relative height of the surface where manipulation is taking place (low, medium, and high). Representative snapshots from one of the videos are shown in Fig. \ref{actions}.
	Each video is annotated manually using the ANVIL annotation tool \cite{kipp2001anvil}. After annotating all the videos, the frames within the same class are extracted and checked for accuracy and consistency. The UW IOM dataset is available for free download and use at: \url{http://dx.doi.org/10.17632/xwzzkxtf9s}.
    
	\section{Experimental Details and Results}
	\subsection{Implementation Details}
	
	For each participant (video), we first compute the REBA score for all the frames. The zero threshold is set to $5^{\circ}$ and the binary threshold is set to $10^{\circ}$. To avoid minor shoulder abductions from contributing substantially owing to Kinect tracking errors, the abduction threshold is chosen as $30^{\circ}$. For the UW IOM dataset, we compute the median of the REBA scores assigned to all the frames belonging to a particular action. We then take the median over all the participants to determine the final REBA score for that action.
	
	The framewise skeletal information available for the TUM Kitchen dataset has a variable lag with respect to the video frames, i.e., the skeleton does not lie exactly on the human pose in the RGB image. Therefore, aggregating over actions and participants according to the RGB image annotations does not result in meaningful REBA scores. We, therefore, reduce the length of both the video annotations of the RGB frames and the framewise REBA scores to 100 using a constant step size of number of frames/100 for every video. We then compute the average REBA score for every action in a particular video using the reduced video annotations and framewise scores. For safety considerations, the maximum score assigned to a particular action among all the videos is considered as the final REBA score for that action.

	The pre-trained VGG16 model for spatial features extraction is trained for 200 epochs with 300 steps per epoch on the TUM Kitchen dataset, and 300 epochs with 300 steps per epoch on the UW IOM dataset with a step-size of $10^{-5}$. The fine-tuned model is trained with the same number of epochs for the TUM Kitchen dataset but with 500 steps per epoch on the UW IOM dataset with 300 steps per epoch and a step-size of $10^{-7}$. The number of training and validation samples for the TUM Kitchen dataset are 24,052 and 5,290, respectively. For the UW IOM dataset, we train over 27,539 samples and validate over 6,052 samples. The models are learned using the 
	TensorFlow machine learning software library \cite{tensorflow} and Python-based Keras \cite{kerasteam2019} neural network library as the backend. To implement the simplified P-CNN model, we modify the MATLAB package provided with \cite{cheron2015p}. Our source code is available at \url{https://github.com/BehnooshParsa/HumanActionRecognition_with_ErgonomicRisk}.
	
	We evaluate the performance of the four segmentation methods by splitting our datasets into five splits, in each of which, the videos are assigned randomly to mutually exclusive training and test sets of fixed sizes. For both the TCN methods, training is terminated after 500 epochs in each of the splits as the validation accuracy stops improving afterward. We use a learning rate of 0.001 for both the methods. D-TCN includes five stacks, each with three layers, and a set of \{32, 64, 96\} filters in each of the three layers. Filter duration duration, defined as the mean segment duration for the shortest class from the training set, is chosen to be 10 seconds. Similarly, training for Bi-LSTM is terminated after 200 epochs for each split as the validation accuracy does not change any further. Bi-LSTM uses Adam optimizer with a learning rate of 0.001, softmax activation function, and categorical cross-entropy loss function.  We choose a linear kernel to train the SVM and use squared hinge loss as the loss function. All the training and testing are done on a workstation running Windows 10 operating system, equipped with a 3.7GHz 8 Core Intel Xeon W-2145 CPU, GPU ZOTAC GeForce GTX 1080 Ti, and 64 GB RAM.

	\subsection{Results}
	\label{sec:Results}
    \subsubsection{Ergonomic Risk Assessment Labels}
    For the TUM Kitchen dataset, fifteen actions are labeled to be medium risk, while the remaining six are deemed as high risk. The high risk actions are associated with closing, opening, and reaching motions, although there is no perfect correspondence due to a lack of fidelity of the skeletal models on which the risk scores are based upon.
    
    In case of the UW IOM dataset, three actions are labeled as low risks, eleven actions are considered medium risk, and the remaining three are identified as high risk. The high risk actions include picking up a box from the top rack and placing objects (box and rod) on the top rack. Walking without holding any object, walking while holding a box, and picking up a rod from the mid-level rack while standing are regarded as low risk, i.e., safe actions. Fig.  \ref{actions} shows the corresponding ergonomic risk labels for these different actions depicted in the video snapshots

    \subsubsection{Video Segmentation Outcomes}
    Table \ref{tum_results} provides a quantitative performance assessment of the two variants of our segmentation method on the TUM Kitchen dataset for camera \# 2 videos. Both the variants perform satisfactorily with respect to all the three performance measures. In fact, the ED-TCN method achieves an F1 overlap score of almost 88\%, which has not been previously reported for any action segmentation problem with more than twenty labels to the best of our knowledge. Our TCN methods also outperform Bi-LSTM and SVM substantially. Just for comparison purposes, it is interesting to note that the pre-trained and fine-tuned VGG16 models provide validation accuracy of 82.80\% and 73.46\%, respectively, during image classification. 
    
    \begin{table}
    \caption{Comparative performance measures of different action segmentation methods on the TUM Kitchen dataset for camera \# 2.}
    \label{tum_results}
    \resizebox{\columnwidth}{!}{%
	\centering
	\begin{tabular}{ ||p{1.5cm}|p{1.5cm}|p{1.6cm}|p{1.7cm}|p{1.7cm}|| }
		\hline
		\multicolumn{2}{|c|}{Method} & Accuracy (\%) & Edit score (\%) & F1 overlap (\%)\\
		\hline
		\hline
		\multirow{4}{1.5cm}{\textbf{Pre-trained VGG16}} & \textbf{D-TCN} & 73.74$\pm$4.57 &78.7$\pm$6.50  & 83.88$\pm$4.52 \\
		\cline{2-5}
		& \textbf{ED-TCN} & \textbf{74.75$\pm$4.08} & \textbf{86.34$\pm$3.15} & \textbf{87.92$\pm$2.16}\\
		\cline{2-5}
		& \textbf{Bi-LSTM} & 62.55 $\pm$ 6.56 & 44.49$\pm$8.67 & 55.11$\pm$9.42\\
		\cline{2-5}
		& \textbf{SVM} & 59.55 $\pm$ 4.98 & 35.39$\pm$3.00 & 47.75$\pm$3.82\\
		\hline
		\hline
        \multirow{4}{1.5cm}{\textbf{Fine-tuned VGG16}} & \textbf{D-TCN} & 74.14$\pm$4.97 &80.33$\pm$5.41  & 84.44$\pm$4.05 \\
		\cline{2-5}
		& \textbf{ED-TCN} & \textbf{74.32$\pm$4.06} & \textbf{84.96$\pm$4.37} & \textbf{87.29$\pm$2.78}\\
		\cline{2-5}
		& \textbf{Bi-LSTM} & 62.89$\pm$ 6.17 & 47.15$\pm$8.67 & 57.75$\pm$9.02\\
		\cline{2-5}
		& \textbf{SVM} & 59.81 $\pm$ 5.10 & 35.8$\pm$3.54 & 47.67$\pm$4.35\\
		\hline
		\hline
	\end{tabular}
	}
    \end{table}

\begin{figure}
        \includegraphics[width=0.5\textwidth]{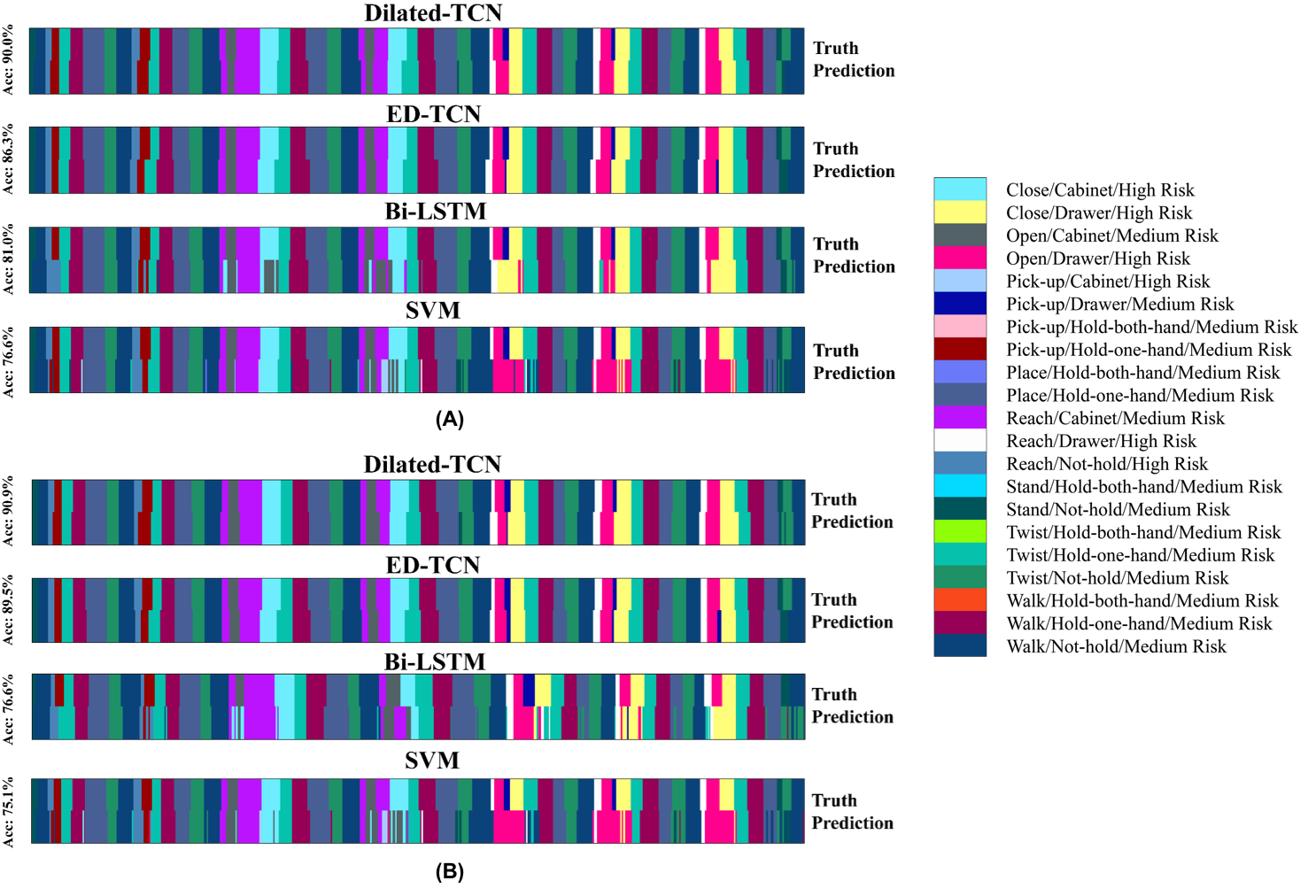}
        \caption{Performance comparison of various methods in action segmentation of a representative TUM Kitchen dataset video using (A) pre-trained VGG16 model and (B) fine-tuned VGG16 model. For each method, the upper row shows the ground truth (manually annotated) action labels, whereas the lower row depicts the corresponding predicted label.}
        \label{tum_segmentation}
    \end{figure}

    Fig. \ref{tum_segmentation} demonstrates that regardless of whether the spatial features are extracted using a pre-trained or fine-tuned VGG16 architecture, both the TCN methods are able to segment the frames into the correct (or more precisely, same as the manually annotated) actions substantially better than Bi-LSTM and SVM. In fact, the global frame-by-frame classification accuracy value is very high, between (86-91)\%, using the TCN methods. Furthermore, both the TCN methods almost always predict the correct sequence of actions unlike the other two widely-used classification methods.
    
\begin{figure}
        \includegraphics[width=0.5\textwidth]{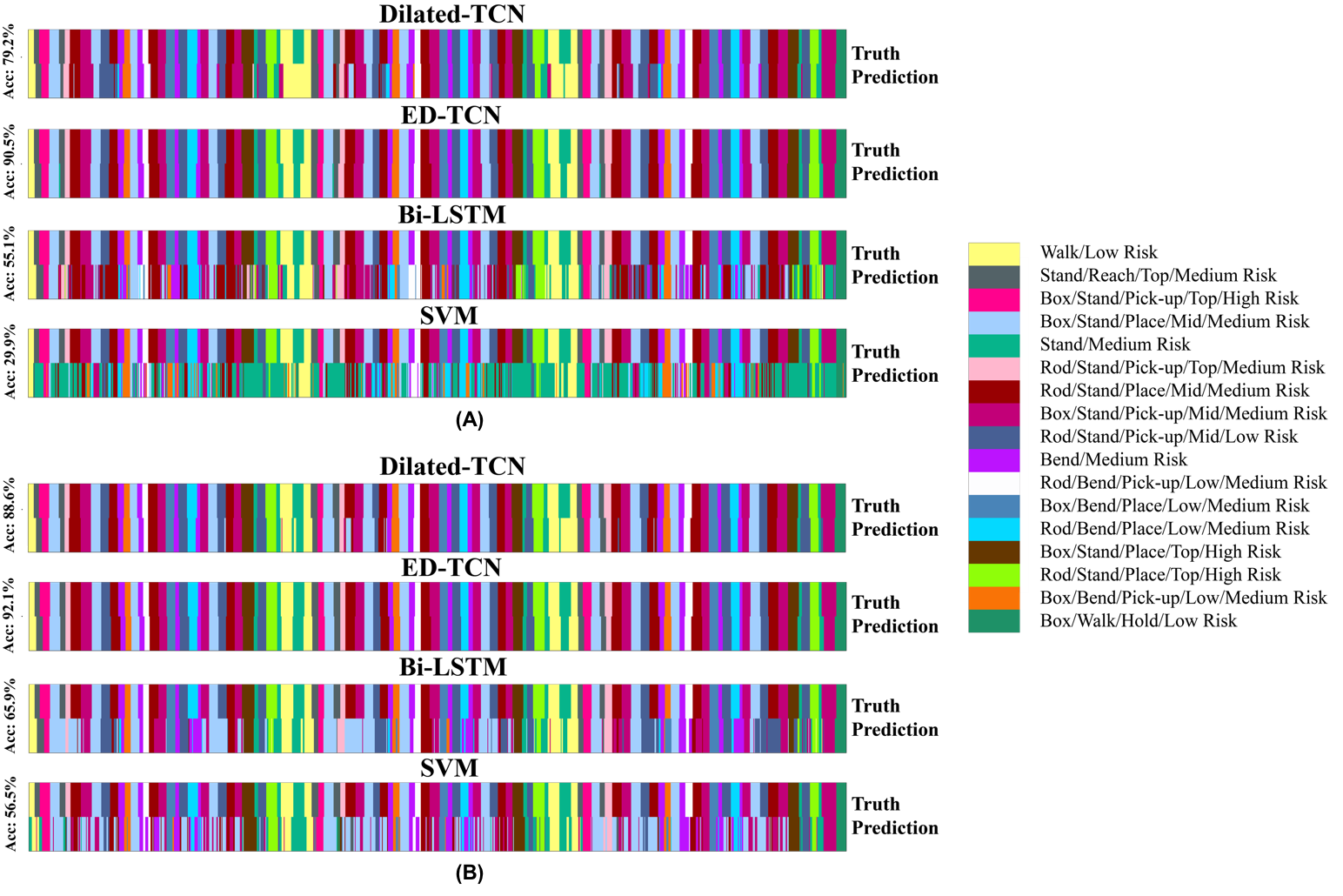}
        \caption{Performance comparison of various methods in semantic segmentation of a representative UW IOM dataset video using (A) pre-trained VGG16 model and (B) fine-tuned VGG16 model. For each method, the upper row shows the ground truth (manually annotated) action labels, whereas the lower row depicts the corresponding predicted label.}
        \label{uw_segmentation}
    \end{figure}

    The difference in performance between the TCN and other two segmentation methods is even more pronounced in case of the UW IOM dataset, which includes a larger variety of object manipulation actions. As shown in Table \ref{UW_results}, SVM performs rather poorly particularly with respect to edit score and F1 overlap values owing to over-segmentation and sequence prediction errors. Bi-LSTM performs somewhat better with the best results obtained using the spatial features generated from a simplified form of P-CNN. Interestingly enough, ED-TCN performs substantially better than D-TCN regardless of the spatial feature extraction method being used. This finding is also consistent with the results for different grocery shopping, gaze tracking, and salad preparation datasets presented in \cite{lea2017temporal}. It happens most likely due to the ability of ED-TCN to identify fine-grained actions without causing over-segmentation by modeling long-term temporal dependencies through max pooling over large time windows. In fact, the edit scores for ED-TCN are close to 90\% and the F1 overlap values are more than 93\% when we use the fine-tuned VGG16 and P-CNN models. The performance measures are almost identical between the two models with P-CNN yielding marginally better results. For just the pre-trained VGG16 and fine-tuned VGG16 models, the validation accuracy is 75.97\% and 73.86\%, respectively, which are similar to the values for the TUM Kitchen dataset. Fig. \ref{uw_segmentation} reinforces these observations on a representative UW IOM dataset video.
    
    \begin{table}
    \caption{Comparative performance measures of different action segmentation methods on the complete UW IOM dataset}
    \label{UW_results}
	\resizebox{\columnwidth}{!}{%
	\centering
	\begin{tabular}{ ||p{1.5cm}|p{1.5cm}|p{1.6cm}|p{1.7cm}|p{1.7cm}|| }
		\hline
		\multicolumn{2}{| c |}{Method} & Accuracy (\%) & Edit score(\%) & F1 overlap (\%)\\
		\hline
		\hline
		\multirow{4}{1.5cm}{\textbf{Pre-trained VGG16}} & \textbf{D-TCN} & 62.11$\pm$4.13 &46.62$\pm$4.17  & 57.48$\pm$5.26 \\
		\cline{2-5}
		& \textbf{ED-TCN} & \textbf{78.76$\pm$3.65} & \textbf{82.96$\pm$3.33} & \textbf{87.77$\pm$2.51}\\
		\cline{2-5}
		& \textbf{Bi-LSTM} & 42.14 $\pm$ 5.45 & 23.76$\pm$1.50 & 29.71$\pm$3.72\\
		\cline{2-5}
		& \textbf{SVM} & 27.10 $\pm$3.40 & 18.05$\pm$0.92 & 20.25$\pm$1.35\\
		\hline
		\hline
        \multirow{4}{1.5cm}{\textbf{Fine-tuned VGG16}} & \textbf{D-TCN} & 61.39$\pm$6.22  & 72.29$\pm$6.16 & 72.29$\pm$6.16\\
		\cline{2-5}
		& \textbf{ED-TCN} & \textbf{86.46$\pm$0.50} & \textbf{88.52$\pm$1.17} & \textbf{93.24$\pm$0.58}\\
		\cline{2-5}
		& \textbf{Bi-LSTM} & 59.23$\pm$4.40 & 33.19$\pm$3.13 & 43.88$\pm$4.23\\
		\cline{2-5}
		& \textbf{SVM} & 42.10$\pm$3.33 & 20.61$\pm$0.89 & 27.56$\pm$1.92\\
		\hline
		\hline
		\multirow{4}{1.5cm}{\textbf{Simplified P-CNN}} & \textbf{D-TCN} & 81.72$\pm$2.82  & 74.01$\pm$5.13 & 82.23$\pm$4.80\\
		\cline{2-5}
		& \textbf{ED-TCN} & \textbf{87.63$\pm$0.77} & \textbf{89.90$\pm$1.16} & \textbf{93.99$\pm$0.77}\\
		\cline{2-5}
		& \textbf{Bi-LSTM} & 71.38$\pm$4.97 & 75.33$\pm$7.41 & 80.45$\pm$7.55\\
		\cline{2-5}
		& \textbf{SVM} & 59.62$\pm$2.74 & 20.09$\pm$0.95 & 31.33$\pm$1.75\\
		\hline
		\hline
	\end{tabular}
    }
    \end{table}

    If we only use the spatial features, image classification validation accuracy is either comparable to (for the TUM Kitchen dataset), or lower than the video segmentation test accuracy (for the UW IOM dataset). Noting that validation accuracy is typically greater than test accuracy for any supervised learning problem, we would expect segmentation accuracy to be much lower than the reported values in the absence of the temporal neural networks. On the other hand, segmentation performance depends quite a bit on the choice of the spatial feature extraction model, particularly in the case of the more challenging UW IOM dataset. This reinforces the intuition that both spatial and temporal characteristics are important in analyzing long-duration human action videos.
    
    It is not surprising to observe that the TCN methods perform better using edit score and F1 overlap score as the measure instead of global accuracy. As also reported in \cite{lea2017temporal}, accuracy is susceptible to erroneous and subjective manual annotation of the video frames, particularly during the transitions from one action to the next, where identifying the exact frame when one action ends and the next one begins is often open to individual interpretation. Both edit score and F1 score are more robust to such annotation issues as compared to accuracy, and, therefore, serve as better indicators of true system performance.
    
    To further evaluate the general applicability of our action segmentation methods, we consider two additional test scenarios: TUM Kitchen videos taken from camera \# 1, and a truncated UW IOM dataset comprising only one sequence of object manipulation actions per participant. Table \ref{tab:Comparative_results} reports the action segmentation outcomes using just the fine-tuned VGG16 model since it yields better results than the pretrained VGG16 model on our regular test datasets. The trends are more or less the same as in our regular datasets. The actual measures are almost identical for the complete and truncated UW IOM dataset. Thus, our methods seem to be robust to sample size, provided all the actions are covered adequately with a sufficient number of instances in the training set, and the actions occur in the same sequence in all the videos. The actual measures for our TCN methods are only slightly lower for the different TUM Kitchen dataset. Thus, the performances appear to be independent of how the videos are recorded. The VGG16 validation accuracy is equal to 76.81\% and 75.28\% for the different TUM Kitchen and the truncated UW IOM dataset, respectively, which are, again, almost identical to the corresponding values for the regular TUM Kitchen and complete UW IOM datasets.

    \begin{table}
    \caption{Comparative performance measures of different action segmentation methods on two additional video datasets.}
    \label{tab:Comparative_results}
    \resizebox{\columnwidth}{!}{%
	\centering
	\begin{tabular}{ ||p{2cm}|p{1.5cm}|p{1.6cm}|p{1.7cm}|p{1.7cm}|| }
		\hline
		\multicolumn{2}{| c |}{Method} & Accuracy (\%) & Edit score (\%) & F1 overlap (\%)\\
		\hline
		\hline
		\multirow{4}{2cm}{\textbf{TUM Kitchen with camera \# 1}} & \textbf{D-TCN} & 64.00$\pm$8.74 & 69.26$\pm$6.08  & 72.24$\pm$7.49 \\
		\cline{2-5}
		& \textbf{ED-TCN} & \textbf{69.83$\pm$4.66} & \textbf{84.69$\pm$3.49} & \textbf{83.43$\pm$3.09}\\
		\cline{2-5}
		& \textbf{Bi-LSTM} & 54.10 $\pm$ 9.37 & 40.64$\pm$8.06 & 48.33$\pm$9.44\\
		\cline{2-5}
		& \textbf{SVM} & 48.43 $\pm$ 8.87 & 30.34$\pm$6.25 & 38.46$\pm$7.88\\
		\hline
		\hline
        \multirow{4}{2cm}{\textbf{UW IOM with non-repeated action sequence}} & \textbf{D-TCN} & 74.04$\pm$2.53 & 62.91$\pm$3.32  & 72.91$\pm$3.01 \\
		\cline{2-5}
		& \textbf{ED-TCN} & \textbf{83.99$\pm$1.10} & \textbf{88.16$\pm$2.24} & \textbf{92.66$\pm$1.72}\\
		\cline{2-5}
		& \textbf{Bi-LSTM} & 58.93$\pm$ 2.22 & 30.57$\pm$2.98 & 41.23$\pm$2.71\\
		\cline{2-5}
		& \textbf{SVM} & 40.62$\pm$1.72 & 21.05$\pm$1.94 & 26.98$\pm$1.92\\
		\hline
		\hline
	\end{tabular}
	}
    \end{table}

    \subsubsection{System Computation Times}
    In addition to characterizing the goodness of action segmentation, we are interested in knowing how long does it take to learn the spatial feature extraction models, to train the segmentation methods, and to compute the framewise action labels during testing. 
    
    The learning times for the pre-trained and fine-tuned VGG16 models are 20,844.11 seconds and 30,564.39 seconds, respectively, in case of the complete UW IOM dataset. As expected, the learning time for the fine-tuned VGG16 model is somewhat lower and equal to 25,414.24 seconds in case of the truncated UW IOM dataset. For the TUM Kitchen dataset, the corresponding value is 31,753.18 seconds.
	
    Using the fine-trained VGG16 model, in case of the complete UW IOM dataset, the overall training times are 252.73 $\pm$ 0.85, 237.76 $\pm$ 0.72, 2,172.23 $\pm$ 11.22, and 60.54 $\pm$ 1.54 seconds across the five data splits for the D-TCN, ED-TCN, Bi-LSTM, and SVM methods, respectively. The corresponding testing times are 0.10, 0.10, 1.09, and 0.09 seconds (the standard errors are negligible), respectively, for an average number of 8,261 frames, which implies that real-time action class prediction is highly feasible. These values are almost identical using the pre-trained VGG16 model. For the TUM Kitchen dataset, the overall training times are 91.19 $\pm$ 1.04, 74.53 $\pm$ 0.65, 619.21 $\pm$ 1.82, and 15.68 $\pm$ 0.67 seconds across the five data splits for the D-TCN, ED-TCN, Bi-LSTM, and SVM methods, respectively. The corresponding testing times are 0.03, 0.02, 0.33, and 0.03 seconds (negligible standard errors), respectively, for an average number of 6,311 frames.
   	
    We further note that the TCN methods also have acceptable training times of the order of a few minutes for reasonably large datasets. This characteristic enables our system to adapt quickly to changing object manipulation tasks. On the other hand, the training times are considerably larger for Bi-LSTM, similar to the results reported in \cite{lea2017temporal}. 
    
    \section{Discussion}
    In case of the more challenging UW IOM dataset, we observe that our TCN methods demonstrate better segmentation performance when spatial features are extracted using the fine-tuned VGG16 model instead of the pre-trained VGG16 model. Consequently, we decided to use P-CNN features to examine whether additional spatial features would further facilitate learning the temporal aspects of the videos for the action segmentation methods. As introduced in \cite{cheron2015p}, P-CNN features are descriptors for video clips that are restricted to only one action per clip. All the frame features of a video clip are aggregated over time using different schemes that result in a single descriptor comprising information about the action in that clip. However, our goal is to process full-length videos with multiple actions. A single time-aggregated descriptor for an entire sequence of multiple actions is not useful to us, as time aggregation results in the loss of important information about the sequence of actions as well as the transitions between the different actions. Hence, we skip the time aggregation step to obtain a video descriptor of the same length as the number of features in the full-length video.
    
    Also, P-CNN features are originally generated by stacking normalized time-aggregated descriptors for ten different patches, i.e., five patches of the RGB image (namely, full body, upper body, left hand, right hand, and full image) and corresponding five patches of the optical flow image. These patches are cropped from the RGB and optical flow frames, respectively, using the relevant body joint positions. The missing parts in the patches are filled with gray pixels, before resizing them as necessary for the CNN input layer. This filling step is done using a scale factor available along with the joint positions for the dataset used in \cite{cheron2015p}. Such a scale factor is, however, not available for our TUM Kitchen and UW IOM datasets. On experimenting with various common values for this scale factor, we observe that it needs to be different for every video as each participant has a somewhat different body structure. Therefore, we only use the full image patches in our simplified form of P-CNN.
    
\begin{figure}
        \includegraphics[width=0.5\textwidth]{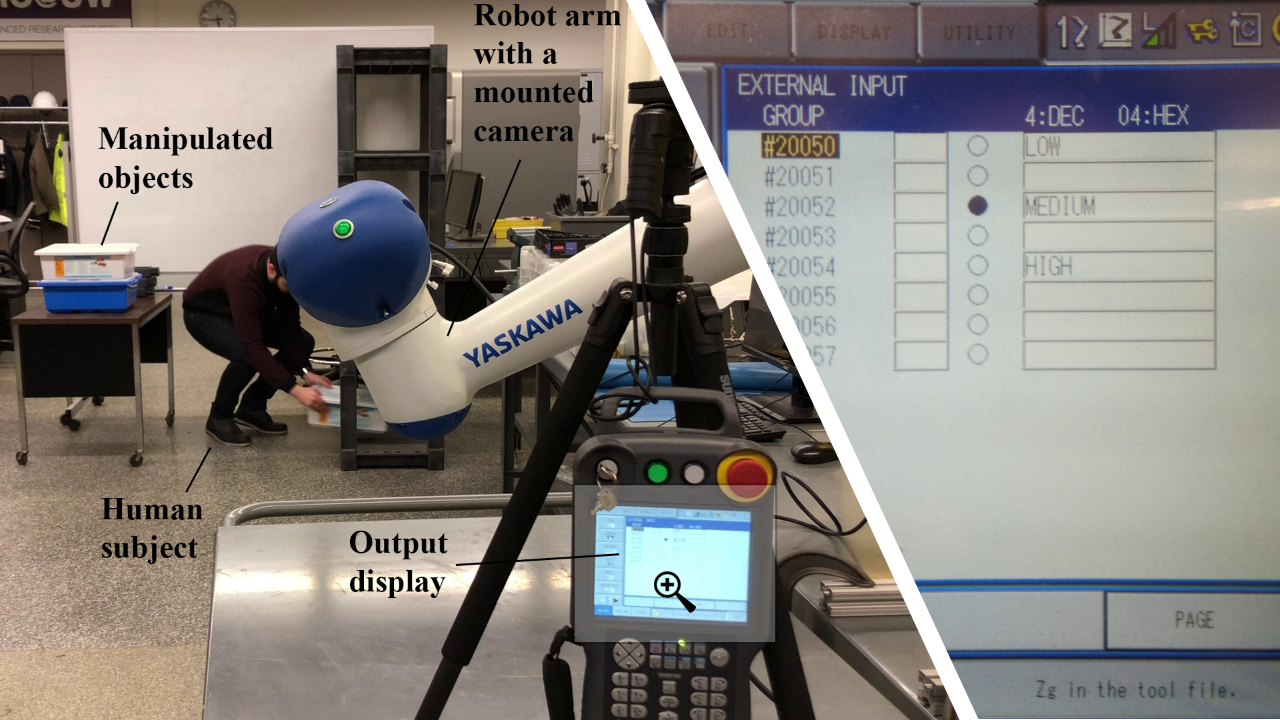}
        \caption{Output indicator corresponding to medium risk is turned on when a human subject performs a Box/Bend/Place/Low action.}
        \label{fig:robotimpl}
    \end{figure}

To further understand the deployment challenges of our system, we perform proof-of-concept trials using a Yaskawa HC10 collaborative robot equipped with an Intel RealSense D435 camera (see Fig. \ref{fig:robotimpl}). Video recordings of eight subjects (four males and four females), each performing the same set of seventeen actions as in the UW IOM dataset, are used to train the segmentation model consisting of the fine-tuned VGG16 and ED-TCN models. The RGB camera frames are captured at 30 Hz and stored using the pyrealsense2 \cite{pyreal} library. At run time, for two new test subjects (one male and one female), the actions are segmented in groups of ninety frames at a time. Considering that ED-TCN requires the features for full-length videos to generate the predictions, we pad the feature vector by repeating the features of the 90\textsuperscript{th} frame. Such padding is done every time ED-TCN is given a new group of ninety frames to segment. The predictions are then communicated to a Programmable Logic Controller (PLC), which displays the outputs (action classes with ergonomic risk levels) on the robot teach pendant. This framework predicts most of the test actions in the correct sequence with reasonably accurate action durations. The performance is somewhat worse than that in Section \ref{sec:Results} due to extra feature padding.

	\section{Conclusions}
	In this letter, we present an end-to-end deep learning system to accurately segment human actions and predict the corresponding ergonomic risks during indoor object manipulation using camera videos. Our system comprises effective spatial features extraction and sequential feeding of the extracted features to temporal neural networks for real-time segmentation into meaningful actions. The segmentation methods work well with just standard (RGB) camera videos, irrespective of how the spatial features are extracted, provided depth cameras are used to generate reliable ergonomic risk scores for all the possible actions corresponding to a known object manipulation environment. Consequently, it makes our system useful for widespread deployment in factories and warehouses without requiring body markers and body-mounted sensors.   
	
	In the future, we intend to further enhance our system to segment the videos satisfactorily, when either the actions are not always performed in the same sequence, or, the same set of actions are not carried out by all the humans. We plan to use the spatiotemporal correlations among the manipulated objects and their affordances, within, potentially, a generative deep learning model, for this purpose. We would also like to develop a learning method that would be capable of risk prediction on a frame-by-frame basis. We then aim to build upon such a method to infer the future actions of a human given a sequence of executed actions, which would be extremely useful in many scenarios. For example, a mobile robot inspector could provide feedback to the workers if they are about to start high risk actions or plan to repeat medium risk actions over extended time periods. Alternatively, the collaborating robot could actively assist in carrying out some of these risky actions. 
	
    \bibliography{references}

\end{document}